\documentclass[letterpaper, 10 pt, conference]{ieeeconf}  

\IEEEoverridecommandlockouts                              

\usepackage{graphicx}
\graphicspath{{images/}}
\usepackage{amsmath}
\usepackage{subcaption}
\usepackage[thinlines]{easytable}
\usepackage{cite}
\usepackage{float}

\usepackage{xcolor}
\usepackage{textcomp}
\usepackage{microtype} 
\usepackage{hyperref}
\usepackage{ccicons}  

\usepackage{xspace}

\usepackage{tabularx}
\usepackage[labelfont=bf,font=footnotesize]{caption}
\usepackage{amssymb}
\usepackage{multirow}
\usepackage{verbatim}
\usepackage{booktabs}

\renewcommand{\baselinestretch}{0.97}

\makeatletter
\DeclareRobustCommand\onedot{\futurelet\@let@token\@onedot}
\def\@onedot{\ifx\@let@token.\else.\null\fi\xspace}

\makeatother
\usepackage{lipsum}

\usepackage{amsfonts}               
\usepackage{mathrsfs}
\let\labelindent\relax
\usepackage{enumitem}              
\usepackage{algorithm}             
\usepackage[noend]{algpseudocode}  

\usepackage[capitalise, noabbrev]{cleveref}

\long\def\xspace{\mathcal{X}}

\newcounter{ocjcount}

\title{\LARGE \bf
{NARF22}: Neural Articulated Radiance Fields \\ for Configuration-Aware Rendering
}

\author{Stanley Lewis \and 
  Jana Pavlasek \and 
  Odest Chadwicke Jenkins
\thanks{%
S. Lewis, J. Pavlasek, and O. C. Jenkins are with the Robotics Institute,
University of Michigan, Ann Arbor, MI, USA,
{\tt \{stanlew, pavlasek, ocj\}@umich.edu}.
The authors would like to thank and acknowledge the work of Stephen Seymour of Morehouse College and Hengxu You, which ultimately led to the contributions contained in this paper.}%
}

\begin{document}

\maketitle
\thispagestyle{empty}
\pagestyle{empty}



\begin{abstract}
Articulated objects pose a unique challenge for robotic perception and manipulation. Their increased number of degrees-of-freedom makes tasks such as localization computationally difficult, while also making the process of real-world dataset collection unscalable. With the aim of addressing these scalability issues, we propose Neural Articulated Radiance Fields (NARF22), a pipeline which uses a fully-differentiable, configuration-parameterized Neural Radiance Field (NeRF) as a means of providing high quality renderings of articulated objects. NARF22 requires no explicit knowledge of the object structure at inference time. 
We propose a two-stage parts-based training mechanism which allows the object rendering models to generalize well across the configuration space even if the underlying training data has as few as one configuration represented.
We demonstrate the efficacy of NARF22 by training configurable renderers on a real-world articulated tool dataset collected via a Fetch mobile manipulation robot. 
We show the applicability of the model to gradient-based inference methods through a configuration estimation and 6 degree-of-freedom pose refinement task.
The project webpage is available at: \href{https://progress.eecs.umich.edu/projects/narf/}{https://progress.eecs.umich.edu/projects/narf/}.
\end{abstract}

\section{Introduction} \label{sec:intro}
Robots operating in environments made for humans must be capable of interacting with a wide variety of articulated objects such as doors, drawers, and hand tools.
In order to perform manipulation tasks involving articulated objects, robots must be capable of localizing the objects and their configurations.
Robust, configuration-aware perception of these objects remains a challenge for robots operating in real-world environments due to the high-dimensionality introduced by articulated degrees-of-freedom and the diversity of object geometries a robot might encounter.

\begin{figure}
    \centering
    \includegraphics[width=\linewidth]{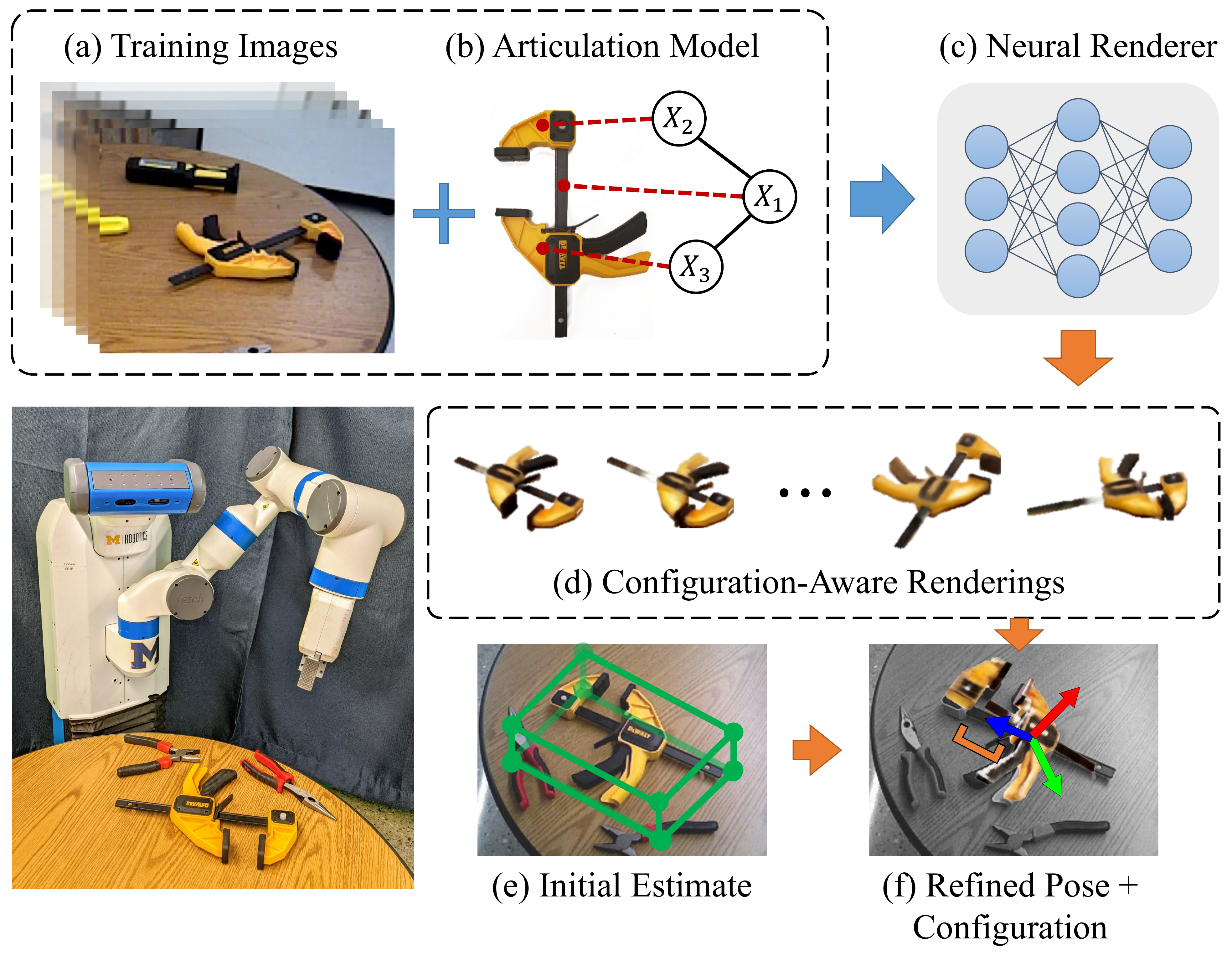}
    \caption{Given robot observations of articulated objects at a small number of configurations (a), along with their poses and articulation models (b), our configuration-aware Neural Articulated Radiance Field (NARF22) method (c) learns to render the object at arbitrary configurations (d). 
    NARF22 renderings of the object at novel configurations and poses can be used to perform pose and configuration estimation of the object (f) given an initial guess (e).}
    \label{fig:nsvf_conf}
\end{figure}

In recent years, data-driven methods have demonstrated impressive results in rigid-body pose estimation from visual data~\cite{xiang2018posecnn, hu2019segmentation, he2021ffb6d}.
These methods have benefited greatly from the availability of standardized real-world datasets~\cite{calli2017yale, hodan2018bop}.
Data-driven \textit{articulated} object estimation is challenging due to the difficulty of generating large-scale datasets which include examples of the full range of configurations. Existing real-world datasets with RGB-D observations and configuration information have been limited to small-scale data collection comprising of a handful of scenes and objects~\cite{martin2018rbo, pavlasek2020parts}.
Following the success of methods for rigid-body pose estimation using synthetic data~\cite{tremblay2018deep}, there has been an emergence of synthetic datasets for articulated objects~\cite{Xiang_2020_SAPIEN, xia2020interactive}. However, the photorealism of the synthetic images is dependent on the quality of the textured mesh models and rendering engine used. Furthermore, scaling such datasets to more object instances and categories may necessitate high-quality textured mesh models, which are expensive to create. 
It stands to reason that being able to generate high-quality articulated object images and meta-data would assist in furthering data-driven approaches to the articulated object pose estimation task.  

Recently, volumetric neural rendering techniques have seen impressive advancements~\cite{mildenhall2020nerf, liu2020neural}.
Neural Radiance Fields (NeRFs) generate highly realistic images of a scene at any given viewpoint by evaluating a differentiable function along rays traced through the scene to obtain colors and densities at each pixel location. 
This technique has the potential to have significant impact on perception for robotics, and has recently shown success when applied to rigid body pose estimation~\cite{yen2020inerf}. The ability to generate photorealistic rendering of an object at any given pose would enable a robot to efficiently evaluate hypotheses about object state against its observations, using the popular render-and-compare strategy~\cite{li2018deepim, deng2019poserbpf}.

This paper introduces a Neural Articulated Radiance Field (NARF22) for creating NeRF style renderers that are paramaterized not only by pose, but additionally by articulated object configuration.\footnote{Note that NARF22 is distinct from NARF~\cite{steder2010narf} for 3 DoF range image feature extraction.}
The configuration-aware neural renderer is trained to render a given articulated object instance using a set of example images of the object with pose and configuration labels
and the object articulation structure in the form of a Universal Robot Description Format (URDF).
We adopt a two-stage pipeline during training. In the first stage, we train individual NeRF style models to render rigid parts of the object. In the second stage, we use the parts-based models and the URDF to augment the training of a single configuration-aware NeRF.
As a result, the method requires only a small number of example configurations in the training data, meaning that fewer manual manipulations of the articulated object are necessary during the real-world data collection stage. Additionally, because the configuration input is part of the NeRF paramaterization, gradient-based optimization techniques can be easily applied with visual or estimated depth supervision. 
To demonstrate the usefulness of this gradient capability, we present a 6 degree-of-freedom (DoF) pose and configuration estimation strategy which leverages the rendering pipeline. This is accomplished through an iterative articulated state hypothesis refinement strategy, inspired by Yen et al.~\cite{yen2020inerf}.

We demonstrate that NARF22 renders photorealistic images of articulated objects at arbitrary poses and configurations when trained on real-world, robot collected data. We further show that the renderings can be used to perform 6 DoF pose and configuration estimation for articulated objects using only RGB observations of a real-world scene.

\begin{figure*}[t]
    \centering
    \includegraphics[width=\linewidth]{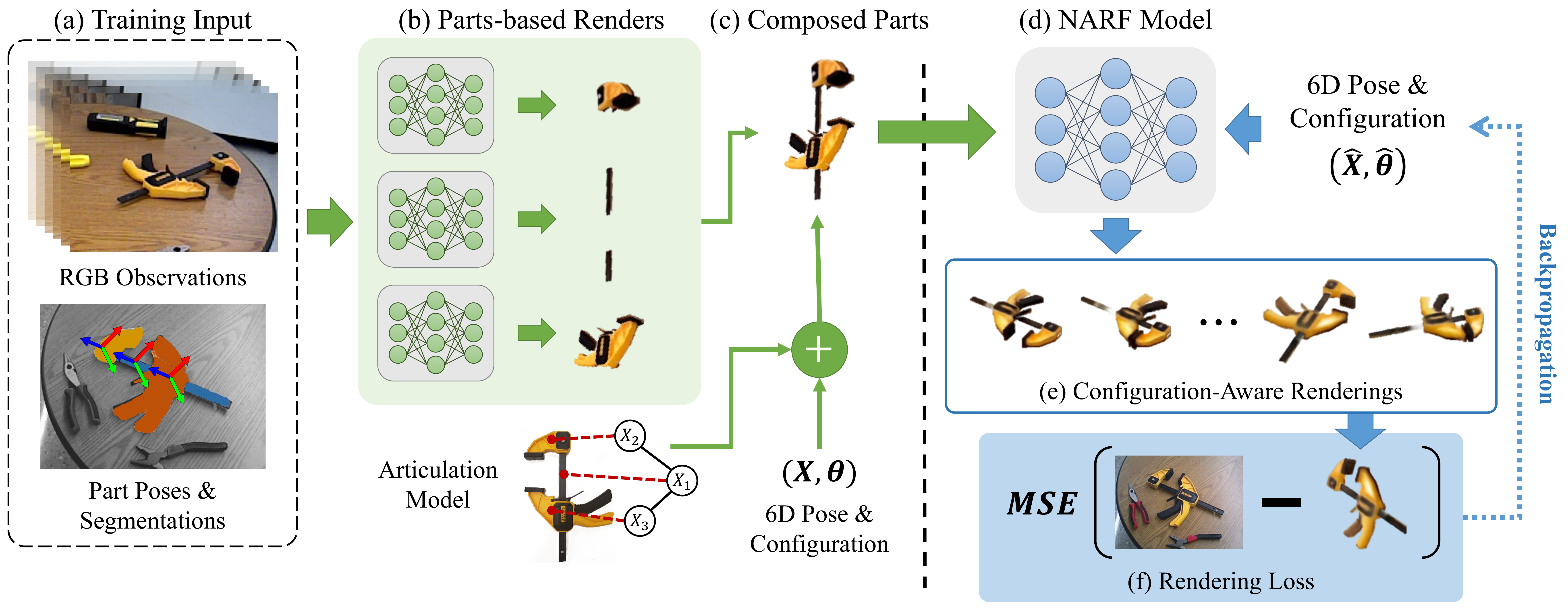}
    \caption{The Neural Articulated Radiance Fields (NARF) pipeline. The training procedure is outlined to the left of the dashed line (with green signals). The inference procedure is on the right (with blue signals). 
    (a) The NARF22 training pipeline takes an input a sequence of images of an articulated object labelled with parts-based poses and segmentations.
    (b) A parts-based neural renderer is trained for each individual object part.
    (c) Using the articulation model of the object, outputs from the parts-based renderers are composed at a given pose $X$ and configuration $\theta$. These composed renderings, along with the state $(X,\theta)$, are used to train the configuration aware renderer, NARF22 (d).
    (e) At inference time, NARF22 can render the object at an arbitrary pose and configuration. 
    (f) The Mean Squared Error (MSE) loss between the rendering and an observation is fully differentiable with respect to a given pose and configuration hypothesis $(\hat{X}, \hat{\theta})$. This enables articulated object pose estimation, similar to Yen et al.~\cite{yen2020inerf}.}
    \label{fig:narf_pipeline}
\end{figure*}

\section{Related Work} \label{sec:related_work}

\textbf{Rigid-body Robotic Perception:}

Recently, data-driven techniques for rigid-body pose estimation have seen success~\cite{xiang2018posecnn, deng2019poserbpf, he2021ffb6d}. A typical commonality is that these methods rely on instance-level information such as textured mesh models or manually labelled datasets~\cite{calli2017yale}. Thus, Marion et al. propose an automatic registration technique dubbed LabelFusion to label poses and segmentation masks for an object across a sequence of images given an initial registration~\cite{marion2018label}, which eases the data generation process. This method assumes that the object is rigid and that the scene is static. Pose estimation has also been accomplished by training on synthetically generated data~\cite{tremblay2018deep}. When researchers posess or can generate high-quality textured mesh models of the relevant objects, this approach works well. These models however may be difficult to obtain, especially if objects are articulated. Large-scale datasets of textureless objects are available~\cite{chang2015shapenet}, but contain little class diversity and have limited application to RGB data. In order to generalize object localization across categories, Manuelli et al. proposed a sparse keypoint-based method and demonstrated its success for robotic manipulation tasks~\cite{manuelli2019kpam}. Such categorical methods are enabled via the scalable data collection methods proposed by Marion et al.

\textbf{Articulated-body Robotic Perception:}

Perception techniques for articulated objects have not benefited from common, large-scale datasets. Instead, articulated object estimation techniques on real-world data have necessitated the laborious generation of custom datasets~\cite{desingh2019pmpnbp, pavlasek2020parts, martin14online, martin2018rbo}. These methods are limited to a small number of object instances, and the addition of new objects requires considerable effort and expertise.
To mitigate the difficulty of manual data collection, several perception techniques have proposed synthetic configuration-aware datasets for point cloud data~\cite{li2020category} and RGB-D data~\cite{xia2020interactive, Xiang_2020_SAPIEN}. These datasets have been used in other recent categorical articulated pose estimation papers such as CAPTRA \cite{weng2021captra}. These methods are highly generalizable but do not result in photorealistic renderings. Zhang et al. propose a method which performs 3D rendering of arbitrary articulated models given a small number of configuration examples, but they do not provide RGB data~\cite{zhang2021strobenet}.

Our proposed method, NARF22, generates photorealistic, configuration-aware RGB-D data which can be used to train perception methods for articulated objects. NARF22 does not require textured models and can be trained using a small number of configurations, which is important as hand labelling real-world data becomes increasingly onerous as dimensions are added to the state space. As presented in this manuscript, NARF22 does rely on a priori knowledge of the object's structure in the form of a URDF file, but works such as Abbatematteo et al., Ajinkya et al., and Yan et al. show that inferring this structure is an active area of research, and thus is left for an area of future investigation \cite{abbatematteo2019learning, jain2020screwnet, yan2019rpm}.

\textbf{Volumetric Rendering:}
Neural rendering has garnered significant attention since the promising results demonstrated by the Neural Radiance Field model (NeRF) by Mildenhall et al.~\cite{mildenhall2020nerf}.\footnote{See also: \href{https://dellaert.github.io/NeRF/}{https://dellaert.github.io/NeRF/}} 
NeRF uses a multi-layered perceptron (MLP) to generate color and density along an epipolar ray as a function of the camera position and orientation. 
This technique is used to render novel viewpoints of a single, static scene, which has the ability to mitigate the burden of hand-annotated training data. However, due to the heavy compute demands of NeRF, several methods have since sought to improve its efficiency. 
Liu et al. improve training efficiency by roughly an order of magnitude by building a voxelization of the scene alongside the model~\cite{liu2020neural}. 
FastNerf, whose network architecture we adopt in this work, adopts a factorization and caching scheme which separates the position and view dependant inputs, thus allowing for improved rendering speed~\cite{garbin2021fastnerf}. Plenoxels is another technique which utilizes direct optimization of plenoptic function parameters volumetrically sampled throughout the scene to dramatically improve both training and inference speed~\cite{yu2021plenoxels}. All of these techniques are however targeted at static scenes with no articulations.

NeRF style models have been shown to be successful on pose estimation tasks by using gradient based optimization on the NeRF model inputs to estimate the pose of a rigid body object~\cite{yen2020inerf}. This result demonstrates that rendered images are of sufficient quality to perform pose estimation, suggesting that such a model could be used in a render-and-compare style of pose estimator.
Other methods have been proposed to extend NeRF models beyond static scenes. Park et al. introduce a deformation model to NeRF which allows for the generation of volumetric representations even with perturbed data, such as hand-held cell phone camera videos of human subjects~\cite{park2020nerfies}. Pumarola et al. utilized a similar deformation model to render dynamic scenes~\cite{pumarola2020d}. 
While these methods handle object deformations, they associate deformations with time or similar latent vector, rather than with an explicit configuration of the object's kinematic model. These models additionally require a well-sampled training distribution across their latent deformation model's inputs, with no method for augmenting poorly sampled training distributions. In contrast, our method extends the FastNeRF model to synthesize a view at arbitrary configurations even if the underlying data has very few configuration values represented. Some works, have also started to consider integrating parts or sub-component based approaches to neural rendering \cite{tseng2022cla,noguchi2021neural}, which lends credence to our approach of parts-informed configuration augmentation and paramaterization.



\section{Background: Neural Radiance Fields}\label{sec:nerf}

Given an arbitrary camera pose, $X$, Neural Radiance Fields (NeRFs) render an RGB image of a known scene through a differentiable function, $F(X)$. 
Traditional NeRF models~\cite{mildenhall2020nerf} are trained by sampling along epipolar rays projected at some subset of image pixels. An encoding of each sample location and ray direction is fed into a multi-layered perceptron (MLP) model, and then integrated along each sampled location on the epipolar ray. This numerical integration process for a ray $r$ is represented as follows:
\begin{align}\label{eqn:volRendTi}
    \hat{C}(r) &= \sum_{i = 1}^N T_i(1-\exp(-\sigma_i\delta_i))\boldsymbol{c}_i \\
    T_i &= \exp \left(-\sum_{j=1}^{i-1}\sigma_j\delta_j \right) 
\end{align}
where $\hat{C}(r)$ is the estimation of the rendered color of the ray, $\sigma_i$ represents the color density value returned from the MLP model for the $i$-th sample, $\boldsymbol{c}_i$ represents the color value returned from the MLP model for the $i$-th sample, and $\delta_i$ represents the Euclidean distance between sample $i$ and its adjacent sample.
A rendered image, $\hat{P}_\text{rgb}$, can be obtained by successive evaluation of $\hat{C}(r)$ for a ray centered at each pixel.


\section{Neural Articulated Radiance Fields}\label{sec:method}


The NARF22 pipeline considers the novel-view, arbitrary-configuration rendering task for articulated objects. 
Given an arbitrary 6 degree-of-freedom (DoF) pose, $X$, and configuration, $\theta$, NARF22 renders an RGB image, $\hat{P}_\text{rgb}$, of the object at the desired state. Configuration represents a vector of joint parameters for each articulating joint in the object. 
The units for each configuration parameter are dictated by the articulation model. For the purposes of this work, we utilize radians for revolute joints and meters for prismatic joints. 
NARF22 seeks a function, $F(X, \theta) = \hat{P}_\text{rgb}$,  which consists of a ray sampling pipeline similar to that described in \cref{sec:nerf}, in which the neural network is conditioned on configuration.


During training, the NARF22 pipeline takes in $N$ images ${\{P_{\text{rgb}}^{(i)}, 1\leq i \leq N\}}$ containing an articulated object along with part-based segmentation masks ${\{P_{\text{mask}}^{(i)}, 1\leq i \leq N\}}$, which contain pixelwise labels for each articulating part. Each frame is labeled with the corresponding ground truth 6~DoF pose of the object in the camera frame and object configuration, $(X^{(i)},\theta^{(i)})$ (see \cref{fig:narf_pipeline}(a)).
The function $F(X, \theta)$ is found by optimizing against a loss function 
over a series of per-pixel rendering operations and comparing against the known ground-truth observed image.

Densely sampling the joint pose and configuration space in real-world collected datasets of articulated objects is challenging due to the need to manually articulate the object and label the corresponding joint values.
To mitigate this, NARF22 utilizes a two-stage training pipeline in which individual rigid parts of the articulated body are first trained without the configuration inputs (see \cref{fig:narf_pipeline}(b)). The parts are then individually rendered and composited into an intermediate dataset representing the full range of possible configurations (see \cref{fig:narf_pipeline}(c)).
In the second stage, a neural renderer is trained with the configuration-parameterized model (see \cref{fig:narf_pipeline}(d)). The training procedure is decribed in more detail in \cref{sec:methods_training}.

The two-stage process allows NARF22 to generalize across articulation configurations when the underlying configuration space is poorly sampled. 
The final model can be rendered at inference time with no knowledge of the articulated object structure. Additionally, gradients can be computed with respect to the configuration vector for a downstream application. We demonstrate this in a pose refinement and configuration estimation task, described below.


\subsection{Pose Refinement and Configuration Estimation}\label{sec:method_pose_est}

We use the configurable NARF22 model, $F(X, \theta)$, to perform configuration estimation and pose refinement from an initial rigid-body 6 DoF pose guess. This method is inspired by Yen et al.~\cite{yen2020inerf}. 
Given an observation $P_{\text{rgb}}$, an initial pose estimate $\hat{X}_0$, and a predicted full-object mask $P_{\text{mask}}$,
a rendered image of the object, $\hat{P}_{\text{rgb}, 0} = F(\hat{X}_0, \hat{\theta}_0)$, is generated. The object pose is relative to the object's root part, as defined by the URDF.
Here $\hat{\theta}_0$ represents our initial configuration guess at step $0$, and is typically set to the midpoint of the allowable joint parameter values.

Since the model is fully differentiable, we can perform iterative gradient-based optimization of the pose and configuration.  
This is accomplished by computing the mean-squared-error loss between the observation values and the rendered image for an estimate $(\hat{X}_k, \hat{\theta}_k)$, at iteration $k$:
\begin{equation}
    \mathcal{L}_{MSE}(\hat{P}_{\text{rgb}, k}) = \big|\big|\hat{P}_{\text{rgb}, k} - (P_{\text{rgb}} * P_{\text{mask}})\big|\big|_2 \label{eq:loss_mse}
\end{equation}
The pose and configuration inputs are updated by computing the gradients of the loss in \cref{eq:loss_mse} to produce new inference-time estimates $(\hat{X}_{k+1}, \hat{\theta}_{k+1})$.
This process is repeated for some user specified $K$ loops, resulting in the final inferred pose and configuration values, $(\hat{X}_K, \hat{\theta}_K)$. This process is visualized in \cref{fig:narf_pipeline}(d-f).


%

\subsection{Training Pipeline}\label{sec:methods_training}

Training the final configuration paramaterized NARF22 model requires a well-sampled configuration space, which for the case of real-world, hand-labeled data, is typically unscalable to generate. Thus, we employ a two-stage pipeline in which we first train a radiance field model on each part of the articulated object which is used to generate augmented training data used to train the final configuration-aware model. 

\textbf{Part-based model training:}
Using training images $P_{\text{rgb}}^{(i)}$, we train rendering models, $F_{j}(X_j^{(i)})$, for each rigid object part $j$. The rendering function is evaluated in the region $P_{\text{rgb}}^{(i)} * P_{\text{mask},j}^{(i)}$ where the $*$ operation denotes the application of the part image mask.
We use the known articulation model to compute part-based poses, $X_{j}^{(i)}$, for a part $j$ in image $i$, in order to train the model.

After training the part rendering models $\{F_j\}$, a new training dataset is generated via a composition of the part models. These compositions are generated by uniformly sampling camera locations on a sphere 1 meter from the object and sampling configuration values uniformly from the allowable configuration space.
We then determine individual part poses from an object pose and configuration using the URDF and render the individual parts at the corresponding locations. Finally, we take the pixels with the smallest predicted depth values from the per-part outputs to generate the final composited tool rendering.
The training procedure is visualized in \cref{fig:narf_pipeline}(a-c).

\textbf{Configuration-aware model training:}
Using the parts-composited data, we train another neural rendering model on the entirety of the tool conditioned on configuration, $F(X, \theta)$. 
An arbitrary neural rendering architecture can be used. The input to the position input portion of the MLP must be modified to be a concatenation of both the pose and configuration vectors.

It is important here to elaborate on why we do not simply stop at compositing the part renderings. Firstly, the final trained model is faster: we observe slightly over a 2$\times$ speedup for the final inference time of the NARF22 model in comparison to the parts-based composition on a desktop workstation equipped with an AMD 5600X CPU and NVidia RTX 3080 GPU. Additionally, the final NARF22 model allows for lower computational requirements during backpropagation, as gradients need only be computed against a single MLP as opposed to one for each part. Furthermore, stopping at only the composition step would require that any downstream task intending to use the part renderers would also need to have access to the URDF structure to obtain part poses from configuration vector and object pose inputs, which 
imposes additional implementation efforts by model end-users. By performing a final model training on this semi-synthetic data, downstream applications require no specialized knowledge of the URDF structure to perform inference. Such knowledge is instead incorporated into the overall NARF22 model. 
We demonstrate how our model can be leveraged in a gradient-based optimization method for configuration estimation and pose refinement by evaluating the gradients directly over the configuration-aware model (see \cref{sec:method_pose_est}).


\section{Experiments} \label{sec:results}


Three primary sets of results are provided in this work. First, qualitative rendering results are presented in which each object's pose and configuration axes are exercised.
Second, quantitative results are presented to demonstrate the rendering accuracy of the proposed configuration-aware NARF22 model versus real-world robot observations. 
Finally, we show results on a configuration estimation and 6 DoF pose refinement task using NARF22 gradients to perform optimization over the estimate.

\subsection{Dataset}
 
Training data is drawn from the Progress Tools dataset used by Pavlasek et al.~\cite{pavlasek2020parts}. This dataset contains approximately 6k RGB-D images at 640x480 resolution taken with a Fetch mobile manipulator robot's onboard Primesense Carmine 1.09 RGB-D sensor. The scenes contain eight hand tools which are divided into their component parts. Mesh models and kinematic models are provided for each object. Per-part and full tool poses and segmentations, as well as configuration parameters, are given for each image.  

\begin{figure}
    \centering
    \includegraphics[width=\linewidth]{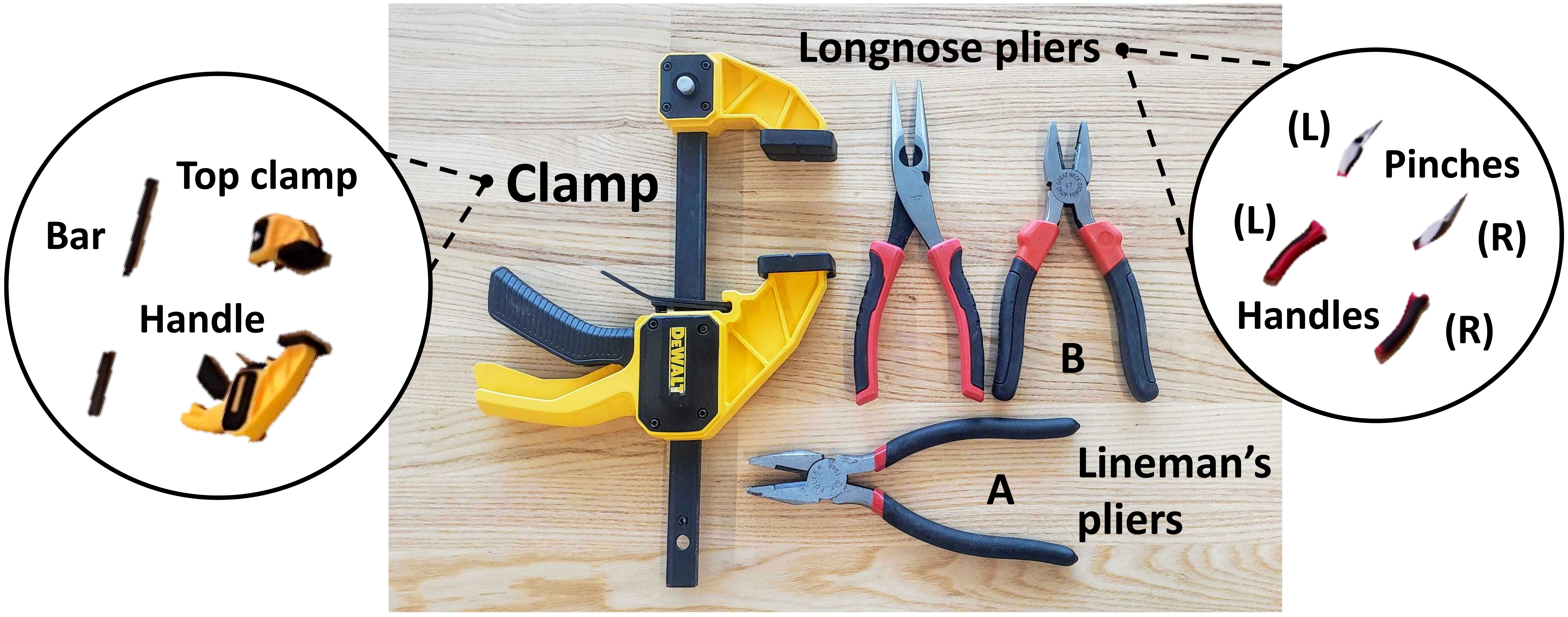}
    \caption{The Progress Tools dataset~\cite{pavlasek2020parts} consists of four articulated hand tools and four rigid body hand tool. This work only utilizes the articulated tools: the clamp, linesman's pliers (A and B) and longnose pliers. The tools are divided into their component parts as shown. All pliers share the same part composition. The parts shown are rendered images from NARF22.}
    \label{fig:progress_tools}
\end{figure}

We focus our attention only on the articulated tools within the dataset: the clamp, lineman's pliers (A and B), and the longnose pliers (see \cref{fig:progress_tools}). The Progress Tools dataset contains both cluttered and uncluttered scenes. We use only the uncluttered scenes of the dataset for training and validation, as handling clutter is outside the scope of this work.
The training data 
contains four configurations of the clamp and one configuration for each of the pliers.
For all experiments, we utilize 5000 intermediate composited training data samples before training the final configurable model.

\subsection{Implementation Details}

\textbf{Model selection:} 
Due to NARF22's nature as a training scheme and MLP paramaterization, it is agnostic to the underlying radiance field MLP architecture. In this work, we present experiments which utilize both Neural Sparse Voxel Field (NSVF)~\cite{liu2020neural} style models, as well as our own NeRF implementation whose MLP architecture is modelled after that of FastNerf~\cite{garbin2021fastnerf}. We modify the input of the models to be the concatenated spatially encoded pose and spatially encoded configuration vectors. For the purposes of this work, the number of spatial encoding levels of the configuration vectors are kept to be identical to that of the position vectors (10 levels). For further details of this spatial encoding process see Mildenhall et al.~\cite{mildenhall2020nerf}.

\textbf{Ray sampling procedure:}
It is important to note that achieving good values of $\delta_i$ for the integration procedure detailed in \cref{eqn:volRendTi} may require a substantial amount of hand tuning with regards to sampling strategy. Mildenhall et al.~\cite{mildenhall2020nerf} solve this problem by including a hierarchical sampling step in which two MLP models are sequentially ran with an importance sampling step in-between. As another example, the Neural Sparse Voxel Field (NSVF) architecture~\cite{liu2020neural} uses a voxel octree to represent the areas of expected epipolar ray returns, and only samples within occupied voxels.

For this work, we present NARF22 pipeline results with an underling NSVF renderer, but we also present an experiment which utilizes a simplified object-centric process containing a pruned axis-aligned voxelized bounding box similar to NSVF, along with a factorized MLP representation analogous to FastNeRF. Our primary NARF22 implementation was trained using consumer grade workstations with single consumer-grade (GeForce) GPUs containing 8~GB of VRAM.

\subsection{Qualitative Results}

\begin{figure*}[t]
      \centering
      \begin{subfigure}[b]{\linewidth}
         \centering
         \includegraphics[width=\textwidth]{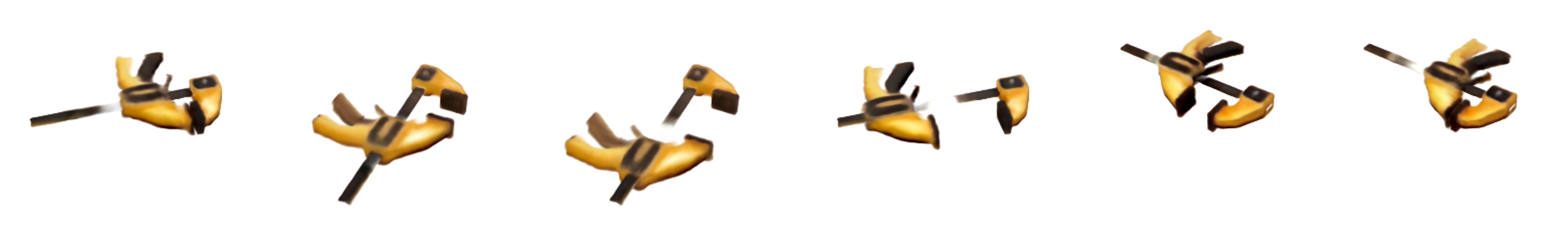}
     \end{subfigure}
     \hfill
     \begin{subfigure}[b]{\linewidth}
         \centering
         \includegraphics[width=\textwidth]{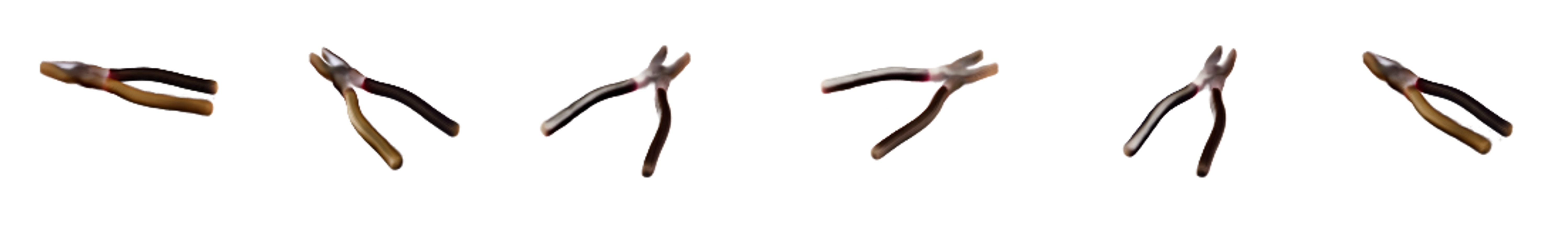}
     \end{subfigure}
     \hfill
     \begin{subfigure}[b]{\linewidth}
         \centering
         \includegraphics[width=\textwidth]{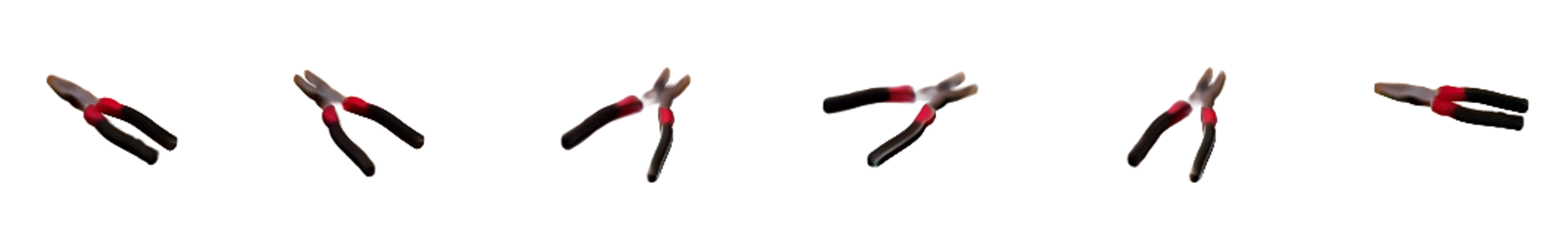}
     \end{subfigure}
     \hfill
     \begin{subfigure}[b]{\linewidth}
         \centering
         \includegraphics[width=\textwidth]{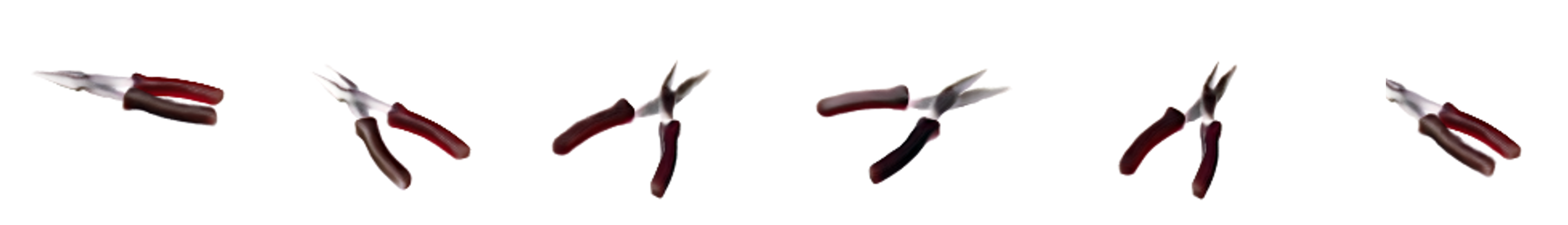}
     \end{subfigure}
    \caption{Rendered images for each tool at novel viewpoints and configurations. From top to bottom: clamp, lineaman's pliers (A), lineman's pliers (B), longnose pliers. 
    }
    \label{fig:qualitative_full_tools}
\end{figure*}

Qualitative results for each of the four articulated tools at a variety of configurations are shown in \cref{fig:qualitative_full_tools}. NARF22 is capable of generating high quality renderings of the tools at arbitrary configurations and viewpoints. We speculate that high amounts of self obscuration of the clamp bar part results in mode collapse during the non-articulated part training process, thus causing irregularities in the final model.  
In contrast, the pliers are able to generalize to novel configurations despite being trained on a single configuration, which we speculate is in part due to their minimal self-occlusions. 

Due to the small size of the pliers in relation to the overall image, particularly with regards to the plier tips, we observe that small labelling errors result in dramatic decrease in final rendering quality. 
The part poses, tool poses, and joint configuration values in our dataset suffer from some degree of error intrinsic to their hand-labelled nature.
These labelling errors, particularly errors in the segmentation masks, negatively affect the quality of the renderings. This phenomenon has been noticed by other researchers as well, such as Abou-Chakra et al.~\cite{abou2022implicit}.  Wang et al.~\cite{wang2021nerfmm} propose a method of dealing with camera pose errors. Incorporating these solutions is left to future work.

Our dataset additionally contains different scenes that do not have precisely identical lighting conditions. Martin-Brualla et al.~\cite{martin2021nerf} proposed a solution to deal with variable lighting, but we consider this to be outside the scope of this work.
Thus, these results include the learned average lighting conditions across the training scenes.

\subsection{Articulated Object Rendering}

Following the proposed parts-based training procedure outlined earlier, the clamp, longnose, and lineman's pliers are evaluated on three collections of different real-world robot collected data. We present the mean squared-error (MSE) between the pixel values for each R,G,B axis (which lie in the range $[0,1]$) in the ground-truth image versus the rendered image in \cref{tab:rendering_metrics}. The `Train' and `Test' datasets consists of images from the uncluttered scenes within the Progress Tools dataset. The individual tool parts were trained using the `Train' dataset whose membership size is shown in Table \ref{tab:rendering_metrics}. The `Test' dataset contains the same scenes and configuration distribution as the `Train' dataset, but with different viewing angles, and thus represents the rendering accuracy of the final full-tool model at poses outside of the training data. The number of tool instances is not equal across the presented metrics because not all tools are present in all images and scenes. 

\begin{figure*}[t]
    \centering
    \includegraphics[width=\linewidth]{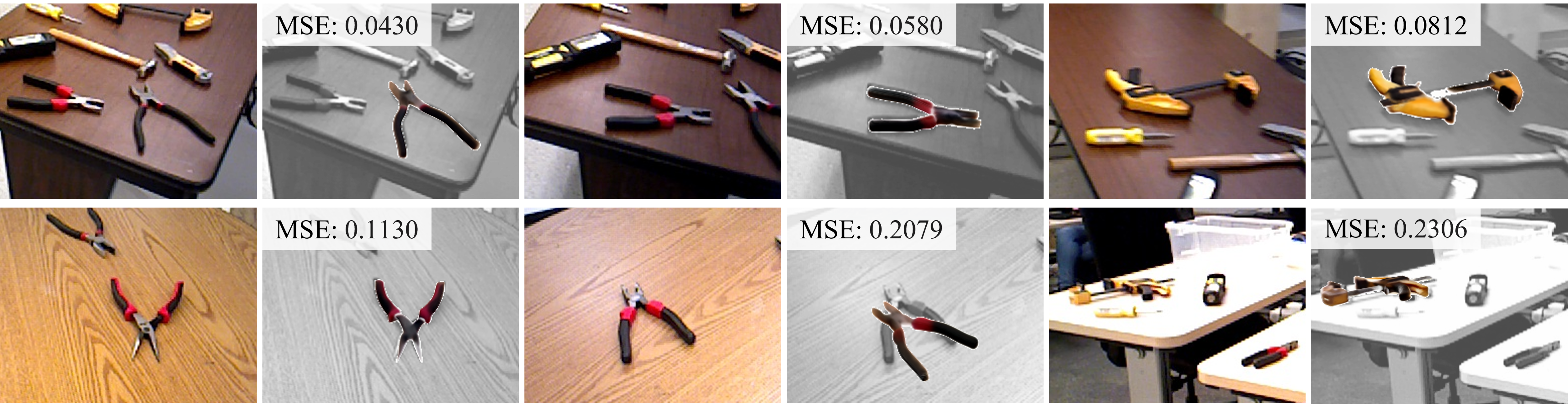}
    \caption{Selected examples of well-performing (top row) and poor-performing (bottom row) renders and their associated MSE. Inaccurate renderings can be caused by difficult viewing angles leading to poor color prediction (bottom right) or occasional mislabeled dataset instances (bottom middle). A typical example of a high MSE rendering is shown in the bottom left. }
    \label{fig:mse_tools}
\end{figure*}

The `Novel Configurations' (Novel Config) dataset consists of additional uncluttered scenes outside of the original training dataset. These scenes contain only samples where the objects are at configurations not present in the training dataset configuration distribution. There are three novel configurations for the clamp and both lineman's pliers, and one novel configuration for the longnose pliers.
Examples of the rendered tools compared to the ground-truth images for the novel configurations are shown in \cref{fig:mse_tools}. 

In general, the pliers are highly sensitive to small errors in ground-truth labelling in the dataset owing to their small geometry. Figure \ref{fig:mse_tools} shows an example of a poor performing plier rendering due to a mislabeled ground-truth mask in the dataset (bottom middle). The pliers also contain only a single, closed configuration within the training set. More configuration examples might improve the results, however qualitative results show that the model generalizes well to novel configurations. The clamp renders are less accurate for extreme poses, such as the one shown in the bottom right of \cref{fig:mse_tools}.

\begin{table}[]
\renewcommand{\arraystretch}{1.1}
\centering
\caption{Render accuracy metrics for the various tools and datasets.}
\begin{tabular}{rlcc}
\toprule
\textbf{Tool} & \textbf{Dataset} & \textbf{Per Pixel MSE} & \textbf{N (instances)} \\ \midrule
\multirow{3}{*}{Clamp} & Test & 0.03343 & 272 \\ 
 & Train & 0.03234 & 2483 \\ 
 & Novel Config & 0.13245 & 200 \\ \midrule
\multirow{3}{*}{Lineman's (A)} & Test & 0.02991 & 171 \\ 
 & Train & 0.02941 & 1557 \\ 
 & Novel Config & 0.10136 & 345 \\ \midrule
\multirow{3}{*}{Lineman's (B)} & Test & 0.06348 & 171 \\ 
 & Train & 0.06318 & 1557 \\ 
 & Novel Config & 0.12500 & 326 \\ \midrule
\multirow{3}{*}{Longnose} & Test & 0.08045 & 171 \\ 
 & Train & 0.07900 & 1557 \\ 
 & Novel Config & 0.10241 & 171 \\ 
\bottomrule
\end{tabular}
\label{tab:rendering_metrics}
\end{table}

\subsection{Configuration Estimation}\label{sec:results_pose}

\begin{figure}[t]
    \centering
    \includegraphics[width=\linewidth]{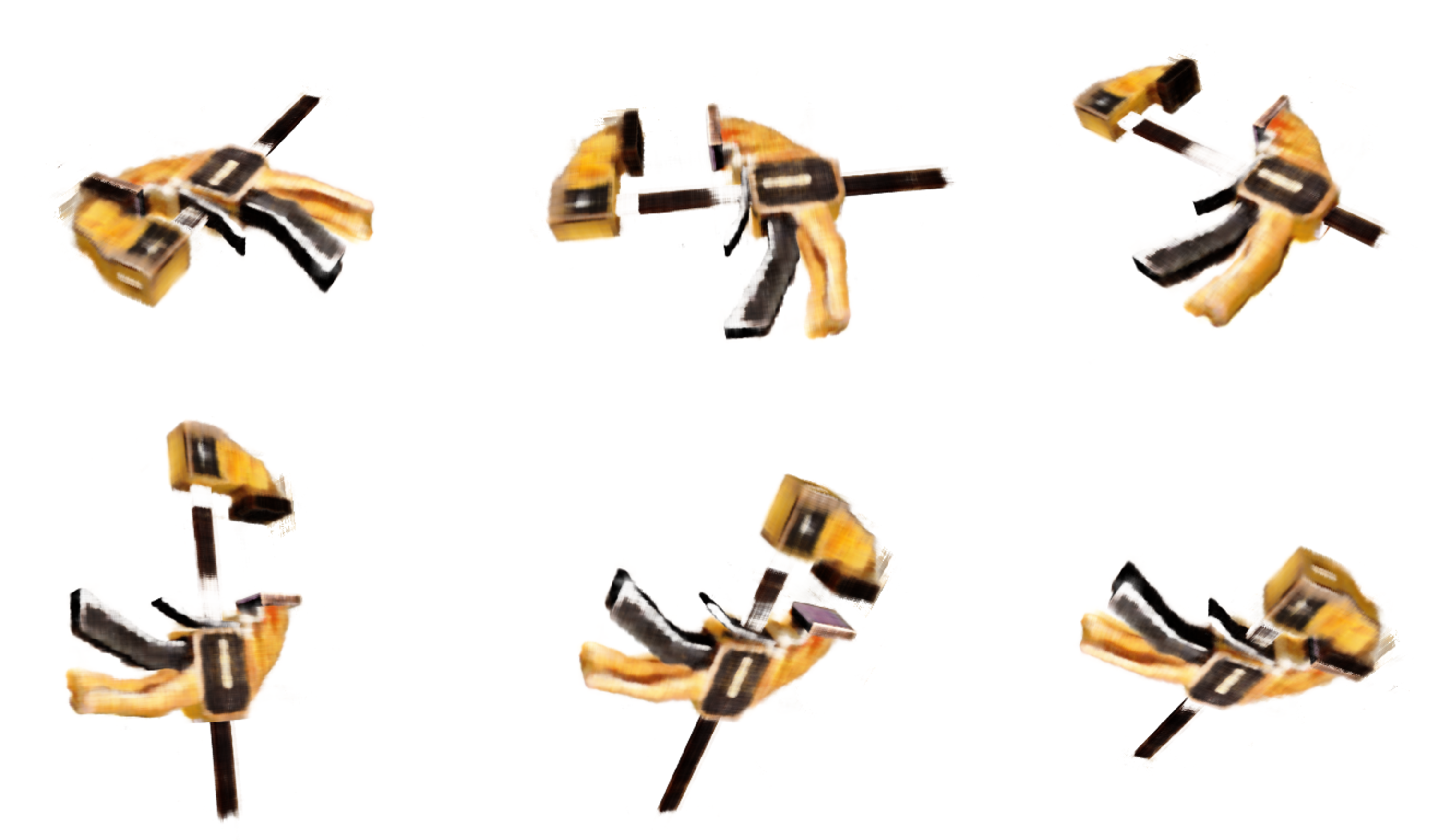}
    \caption{Qualitative renderings of the clamp model used for configuration estimation results. These renderings come from the FastNeRF style of factorized MLP paired with the voxelized sampling mechanism.
    }
    \label{fig:clamp_fastnerf}
\end{figure}

\begin{figure}[t]
    \centering
    \includegraphics[width=\linewidth]{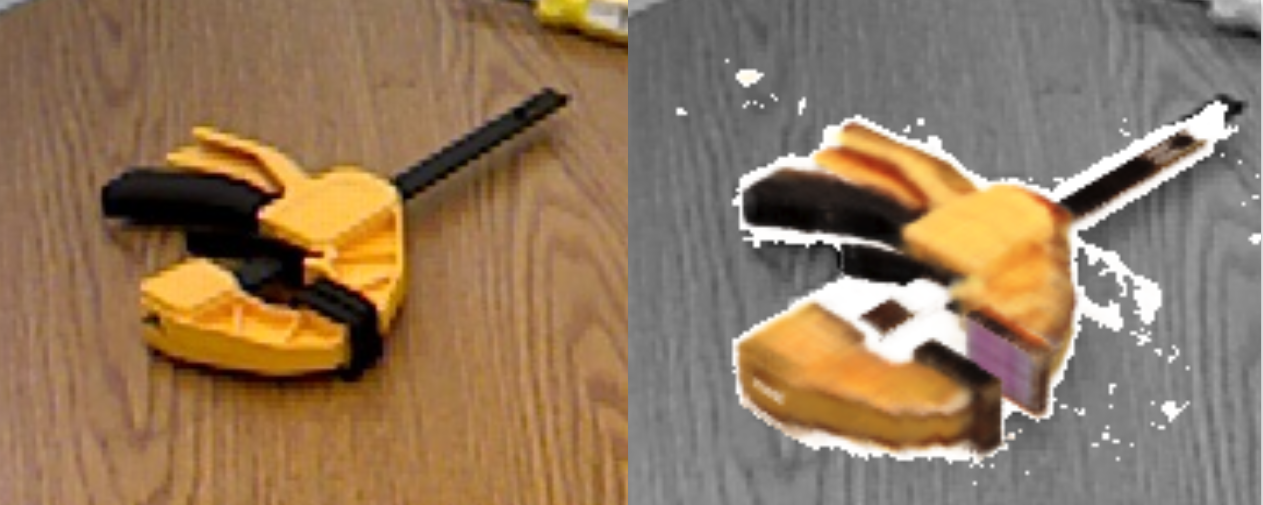}
    \caption{On the right, the worst configuration estimate rendering is overlaid onto the ground-truth via the segmentation mask. On the left is the original observed image. Images are cropped and zoomed for clarity.}
    \label{fig:worstRender}
\end{figure}

For a final demonstration, we perform pose refinement and configuration estimation from an initial rigid-body pose estimation. We train a NARF22 model of the clamp in which the underlying MLP is our own implementation of the FastNeRF style architecture, with the clamp configuration vector added to the position-dependant MLP. Unlike the original FastNeRF as proposed by Garbin et al.~\cite{garbin2021fastnerf}, we do not utilize any kind of caching or other speedup mechanisms. Some qualitative renderings of the clamp in this model are shown in Figure \ref{fig:clamp_fastnerf}. 

After training, we select 17 uncluttered scenes from the test set of the Progress Tools dataset~\cite{pavlasek2020parts}, and initialize the pose of our configurable model at the ground-truth pose of the clamp handle, but with 10 degrees of uniformly sampled rotation pertubation in azimuth and elevation, and 2~cm of uniformly sampled perturbation in translation. The configuration is initialized to be the center of the allowed configuration range.
We then perform 150 iterations of gradient descent optimization on the joint configuration and pose inputs to our renderer using an Adam optimizer in which the learning rate for the configuration parameters is set to 0.01, and to 0.001 for the pose parameters. This style of estimation is similar to Yen-Chen et al.~\cite{yen2020inerf}, however this time we also include a configuration input on which we have no prior information.
This procedure is analogous to performing articulated body pose estimation by starting with a rigid body pose estimate of the most prominent part of the articulated body (e.g. from PoseCNN~\cite{xiang2018posecnn} or a similar estimator). 

We recover the configuration estimation extremely well, with all test samples except 3 achieving sub-centimeter accuracy on the clamp's prismatic joint parameter, and only one exceeding 2 cm of error. In Table~\ref{tab:config_est_metrics} we report summary statistics of configuration error, along with a full-tool matching score in the form of an average distance (ADD) loss metric between the full object pose and configuration ground-truth and our estimation. ADD measures the average Euclidean distance between corresponding points on the model of the configurable object when transformed to both the ground-truth and estimated pose and configurations. For the full specification of the ADD loss metric, see Xiang et al.~\cite{xiang2018posecnn}. The worst configuration estimation is shown alongside the ground-truth in Figure \ref{fig:worstRender}.

\begin{table}[]
\renewcommand{\arraystretch}{1.1}
\centering
\caption{Summary statistics for the configuration estimation experiment.}
\begin{tabular}{ccc}
\toprule
\textbf{Metric} & \textbf{Mean} & \textbf{Std. Dev.} \\
\midrule
ADD (m)                      & 0.0107            & 0.0043                 \\
Configuration Err. (m)       & 0.0073            & 0.0065                 \\ \bottomrule
\end{tabular}
\label{tab:config_est_metrics}
\end{table}

\section{Conclusion} \label{sec:conclusion}
In conclusion, a training pipeline which is capable of including the configuration vector of a real-world tool and subsequently produce arbitrary view and configuration renderings is presented. This output is enabled by a parts-based two-stage training process. By utilizing information about tool structure, a semi-synthetic set of augmented data is generated, which then is used to create a single full-object renderer. This final renderer is demonstrated to be agnostic to the underlying neural rendering architecture by providing quantitative results from an NSVF based renderer, and providing configuration estimation results from a FastNeRF inspired MLP renderer. The utility of being able to perform gradient descent to the join position and configuration inputs is demonstrating by showing accurate results.


\bibliographystyle{IEEEtran}
\bibliography{ref.bib}

\end{document}